\documentclass[11pt]{article}
\usepackage{acl2016}
\usepackage{times}
\usepackage{latexsym}

\usepackage[explicit]{titlesec}
\usepackage{booktabs}
\usepackage{graphicx}
\usepackage{float}
\usepackage{color}
\usepackage{amssymb}
\usepackage{mathtools}
\usepackage{multirow}

\aclfinalcopy 

\title{Multilingual Modal Sense Classification using a Convolutional Neural Network}

\author{Ana Marasovi\'c \and Anette Frank \\
	    Research Training Group AIPHES\\
	    Department of Computational Linguistics\\
	    Heidelberg University\\
	    69120 Heidelberg, Germany\\
	     {\tt \{marasovic,frank\}@cl.uni-heidelberg.de}
}

\date{}

\begin{document}


\clearpage\maketitle
\thispagestyle{empty}

\begin{abstract}
Modal sense classification (MSC) is a special WSD task that depends on the meaning of the proposition in the modal's scope. We explore a CNN architecture for classifying modal sense in English and German. We show that CNNs are superior to manually designed feature-based classifiers and a standard NN classifier. We  analyze the feature maps learned by the CNN and identify known and previously unattested linguistic features. We benchmark the CNN on a standard WSD task, where it compares favorably to models using sense-disambiguated target vectors. 
\end{abstract}

\section{Introduction}

Factuality recognition \cite{Marneffeetal:12,lee-EtAl:2015:EMNLP}
is a subtask in information extraction that differentiates
facts from hypotheses and speculation, expressed through signals of
modality, most prominently, modal verbs and adverbs.  
Modal verbs are, however, ambiguous between an {\em
  epistemic} sense (possibility) as opposed to non-epistemic {\em
  deontic} (permission/obligation) or {\em dynamic} (capability)
senses, as in: {\em He could be at home} (epistemic), {\em You
  can enter now} (deontic) and {\em Only John can solve this problem} (capability).

Modal sense classification (MSC) is a special case of sense disambiguation
that is also relevant in areas of dialogue act and plan recognition
in AI, as well as novel tasks such as argumentation mining. Prior work
\cite{Ruppenhofer2012,Zhouetal:15}
addressed the task with feature-based classification. However, even
with carefully designed semantic features the models have difficulties beating the majority sense
baseline in cases of difficult sense distinctions and when
applying the models to heterogenous text genres.
 
We cast 
modal sense classification as a novel semantic sentence classification task using a convolutional neural network (CNN) architecture.
Our contributions are:
(i) our experiments on MSC confirm the adequacy of CNNs for
modeling propositions in {\em semantic} sentence
classification tasks (cf.\ 
\newcite{kim:14});
(ii) we show that automatically learned features
in a CNN outperform manually designed features 
for difficult modal verbs and novel genres; (iii) we demonstrate that the CNN
approach can be generalized across languages, by adapting the
model to German. (iv) We offer insights into the linguistic
properties captured by the learned feature maps. 
Finally, (v) we benchmark the CNN on a standard WSD task, comparing it to a WSD model using rich sense-disambiguated embeddings and obtain comparable results.

\section{Prior and related work}

\paragraph{Modal sense classification (MSC).} 
We focus on disambiguation of modal verbs,
adopting the sense
inventory established in formal semantics: {\em epistemic,
  deontic/bouletic} and {\em circumstantial/dynamic}.\footnote{These senses correspond to \cite{Bakeretal:2010}'s
modal categories (with {\em deontic} split into {\em requirement} and
{\em permissive}), and R\&R’s inventory, with regrouping of {\em concessive, conditional}
and {\em circumstantial}, cf.\ \newcite{Zhouetal:15}.} 
We compare to prior work in \newcite{Ruppenhofer2012} and follow-up work
in \newcite{Zhouetal:15} (henceforth, R\&R and Z+).
R\&R induced modal sense classifiers from manual annotations 
on the MPQA
corpus \cite{Wiebeetal:05} using 
word-based and syntactic
features. Z+ propose an extended semantically informed
model that significantly outperforms R\&R's
results.
Z+ also create heuristically
sense-annotated training data from parallel corpora, to overcome sparsity and bias
in the MPQA corpus. However, their models do not beat the
majority sense baseline for the difficult modal verbs,
{\em may, can} and {\em could}.

Modal sense classification  interacts with genre and
domain differences. \newcite{prabhakaran-EtAl:2012:ExProM} 
observe strong cross-genre effects and missing generalization
capacities when applying their modality classifier to out-of-domain genres.

\paragraph{Word Embeddings and Sense Disambiguation.} 

\newcite{taghipour2015semi} investigate the impact of word embeddings
on classical WSD, using pre-trained embeddings and tuning them to the task using a
NN. Both variants, integrated into the state-of-the-art
system IMS \cite{zhong-ng:2012:ACL2012},  improve WSD performance
on benchmark tasks.

Ordinary word embeddings
do not differentiate word senses. \newcite{rothe-schutze:2015:ACL-IJCNLP}
explore supervised WSD using
sense-specific embeddings, which they induce
by exploiting {\em sense encodings} and {\em constraints}
given by a lexical resource.\footnote{ 
Modal verbs are not or not
systematically covered in WordNets or VerbNet;
FrameNet relates modal verbs to their
predominant sense only. Also,
FrameNet's 
frame-to-frame relations are known to lack coverage \cite{Burchardtetal:09}.}
Integrating the sense-specific
vectors into IMS 
yields significant improvements 
and small gains relative to
\newcite{taghipour2015semi}.
Hence, word embeddings -- tuned to the task or sense-specific -- prove
beneficial for supervised WSD.


The CNN approach we investigate in our work 
does not employ a fixed feature space or a pre-defined window around the target word. It
flexibly learns feature maps for variable window sizes over the embedding matrix for the full sentence. In contrast to \newcite{rothe-schutze:2015:ACL-IJCNLP}, embeddings used by our CNNs models are knowledge-lean and do not encode senses of the target words.


\if false
Semi:
- PoS, context lexical, collocation features
- fixed window size + target word
(MSC: fixed words and syntactic position; 
-> ignore POS, capitalization
-> instead of tuning window size -> use CNN for leaning useful feature
maps of different window sizes)
NN uses POS and Capitalization

(\newcite{2015arXiv151106388T})
\fi

\paragraph{Sentence classification using CNNs.}
Recent work investigates NN architectures and
their ability to capture the semantics of sentences for various
classification tasks. \newcite{kalchbrenner-grefenstette-blunsom:2014:P14-1} construct a
dynamic CNN that builds on unparsed input and achieves
performance beyond strong baselines for sentiment and
question type classification. By contrast, recursive neural networks
\cite{socher-EtAl:2013:EMNLP} take parsed input, 
recursively generate representations for intermediate phrases,
and perform classification on the basis of the full
sentence representation.

\newcite{kim:14} evaluates a one-layer CNN on various benchmark tasks for sentence classification.
CNNs trained on pre-trained (static) embeddings perform well and can be further improved by tuning them to
the task
 (non-static). Using two channels did not significantly improve results.
Overall, the CNNs
show consistently strong performance, improving on state-of-the-art results in 4 out of 7 tasks, i.a.,\ sentiment and opinion classification.

\section{A CNN for modal sense classification}

We aim at a NN approach to MSC that (i) improves over existing feature-based classifiers, (ii) alleviates manual crafting of features, (iii) generalizes over various text genres, and (iv) is easily portable to novel languages.
Besides this, MSC
is a special kind of WSD, in that modal verbs
have a restricted sense inventory shared across languages, and 
act as operators that take a full proposition as argument. 
We thus cast MSC as a semantic sentence classification task 
in a CNN architecture, adopting the one-layer CNN model of \newcite{kim:14}, a variant of  \newcite{collobert2011natural}. Unlike \newcite{kim:14} we will use only one channel, but experiment with various types of word vectors.

A CNN represents a sentence with a fixed size vector, passed to classifier to classify the sentence into task-specific target categories. In our case, it will classify sentences into three modal sense categories. The input layer is a matrix ${\bf x} \in \mathbb{R}^{s \times d}$, with each row corresponding to a $d$-dimensional word embedding $x_{i} \in \mathbb{R}^{d}$ of a word in the sentence of length $s$. Word embeddings can be randomly initialized or pre-trained vectors, e.g.\ word2vec \cite{mikolov2013distributed} or dependency-based \cite{levy2014dependency} embeddings. Based on the input layer, a CNN builds up one or more convolutional layers. A convolution is an operation between sub-matrices of the input matrix ${\bf x} \in \mathbb{R}^{s \times d}$ and a {\em filter} parametrised by a weight matrix ${\bf w} \in \mathbb{R}^{n \times d}$, that returns a vector usually referred to as a {\em feature map}. Formally, let ${\bf x}_{i-n+1:i}$ be the sub-matrix of the input matrix ${\bf x}$ from the $(i-n+1)$-th row to the $i$-th row and let $\langle . \,, . \rangle_{\mathcal{F}}$ denote the sum of elements of the component-wise inner product of two matrices, known as Frobenius inner product. The $i$-th component of the feature map {\bf c} is obtained by taking the Frobenius inner product of the sub-matrix ${\bf x}_{i-n+1:i}$ with the filter matrix ${\bf w}$
\begin{equation}
    c_{i} = \langle {\bf x}_{i-n+1:i}{,}{\bf w} \rangle_{\mathcal{F}}\,,
\end{equation}
for $i \in \{n, \hdots, s\}$\footnote{We apply the narrow type of convolution.}. Afterwards, we add a bias term, $b \in \mathbb{R}$ to every component of the feature map and apply an activation function $f$,
\begin{equation}
    \tilde{c}_{i} = f(c_{i} + b)\,.
\end{equation}
Finally, {\em max-over-time pooling} \cite{collobert2011natural} is applied over a single feature map that extracts the maximum value $\hat{c} = max \{{\bf \tilde{c}}\}$, which represents the chosen feature for this feature map. Like \newcite{kim:14} we don't use just one filter as described, but multiple filters with different region sizes $n$, resulting in multiple feature maps. Features obtained through max-pooling from each feature map are concatenated to a 
vector representation of the input sentence that is passed to the softmax layer. Parameters to learn are elements of the filter matrices and the input matrix when word vectors are tuned. 

Filters are trained to be especially active when they encounter a sequence of words relevant for the given classification task. 
\newcite{kalchbrenner-grefenstette-blunsom:2014:P14-1} present $n$-grams of different feature detectors that capture positive or negative sentiment phrases, and also more abstract semantic categories, such as negation or degree particles ('too') that are relevant in compositional sentiment detection. 
In the modal sense classification task, we expect the feature maps to capture semantic categories found to be relevant in prior work, such as tense, aspectual classes, negation and semantic properties of verbs and phrases. Moreover, prior work
has shown that MSC profits from features that model the wider syntactic context, esp.\ subject and embedded verb and their semantics (abstractness, semantic class, aspect, tense).
Explicit modeling of these features as in Z+ improves performance, but requires 
feature design for each new language. 
Also, modeling semantic features through 
lexical resources is subject to sparsity, and relying on parsed input leads to lack of robustness. 

Given that MSC 
profits from semantic features in the wider syntactic context, we expect that a CNN that applies filters of variable sizes 
to various regions of the sentence to learn feature maps can capture diverse linguistic features, and
offers greater flexibility compared to a conventional WSD model with a fixed window size centered around the target word.
To investigate these special properties of the CNN model, 
we test it on English and German data. While in English, subject, modal and embedded verb are in a close syntactic context, in German, they can be distributed over wider distances, and the feature maps are expected to capture properties over wider distances.

We perform experiments for MSC for English and German, using various data sets. Section 4 presents the data, experimental settings and the model variations we investigate. We perform detailed quantitative and qualitative evaluation of our experimental results.
In Section 5, we evaluate the CNN approach in a lexical sample WSD task, to benchmark its performance on a well-studied data set, and to investigate the potential advantage of learning feature maps based on flexible window sizes. To our knowledge, this constitutes the first attempt to apply a CNN model in a WSD task.



\section{Modal sense classification}

\subsection{Data}
Our experiments are based on three data sets. Their basic composition is given in Table \ref{tbl:datasets}.\footnote{More detailed information will be provided through accompanying material with the final version. The  annotated MASC and EPOS$_G$ data sets will be made publicly available.}

\paragraph{1) MPQA + EPOS$_E$} The English benchmark data set MPQA from R\&R was further enriched through balanced heuristically tagged training data, EPOS$_E$, by Z+. The EPOS$_E$ data set was obtained  using a cross-lingual sense projection approach.
Z+ identified paraphrases for modal senses (e.g.\ {\em brauchen-need; erlauben-permit } for deontic, {\em schaffen-able to} for dynamic sense), extracted sentences from a parallel corpus with a modal verb aligned to a sense-identifying paraphrase, and tagged them with the identified modal sense. Z+ measured 0.92 accuracy on 420 instances of the heuristically tagged corpora. To alleviate distributional bias stemming from the MPQA dataset, Z+ balanced the blend of MPQA with EPOS$_E$ using under- and oversampling. We experiment with both versions ($\pm$ balanced).\footnote{Their data is publicly available through their website. We omit {\em shall} from MPQA, due to low number of occurrences.}

\paragraph{2) MASC} A subset of the multi-genre corpus MASC \cite{masc2008}, consisting of 19 genres was manually annotated (Anonymous) with modal senses for the same modal verbs. The annotated data consists of  $\approx$100 instances for each genre.\footnote{Exceptions with less than 100 instances are journal, newspaper, technical, travel guides, and telephone.}

\paragraph{3) EPOS$_G$} Following the method of Z+, we constructed a German data set EPOS$_G$ from the Europarl and OpenSubtitles corpora of OPUS \cite{TIEDEMANN12.463.L12-1246} by projecting modal sense categories from English to German, using selected modal sense identifying English paraphrases. 
The resulting corpus with sense-tagged German modal verbs {\em k\"onnen (can), m\"ussen (must), sollen (should), d\"urfen (may)}
consists of a manually validated test section consisting of up to 100 instances for each sense. Annotation was done by two independent judges and one adjudicator.
Balanced training data of 1000 instances per sense for each modal verb was constructed from heuristically tagged sentences that were judged high-quality by validating 20 instances for each paraphrase. For modal verbs with rare extractions, we added training data from modal verbs of shared senses, changing their verb forms to the verb form of the target verb.\footnote{Replacing e.g.\ {\em k\"onnte} with {\em d\"urfte} in {\em Es k\"onnte Dir gefallen} extracted from {\em You might get a taste for it.}}

\begin{table}
\small
\center

\resizebox{0.40\textwidth}{!}{
\begin{tabular}{c|c|ccccc}
\toprule
\multicolumn{1}{r}{ } & \multicolumn{1}{r}{ }& \multicolumn{1}{r}{can}& \multicolumn{1}{r}{could} & \multicolumn{1}{r}{may} & \multicolumn{1}{r}{must} & \multicolumn{1}{r}{should}\\\midrule
\multirow{3}{*}{MPQA}	 & ep &   2 & 156 &  130   &  11  &      26\\
 & de & 115   & 17         &  9   &    83      &      248\\
 & dy & 271   & 67         &  --   &    --    &      --\\
 \midrule
\multirow{3}{*}{EPOS$_E$}	 & ep & 150   & 40         &  950   &   800      &      150\\
 & de & 150   & 40         &  950   &    800      &      150\\
 & dy & 150   & 40         &  --   &    --      &      --\\
 \midrule
\multirow{3}{*}{MASC}	 & ep & 88   & 144         & 217   &    29      &      27\\
 & de & 72   & 16         &  43   &    115      &      224\\
 & dy & 710   & 251         &  3   &    --     &      --\\ 
\bottomrule
\end{tabular}
}
\resizebox{0.40\textwidth}{!}{
\begin{tabular}{c|c|cccc}
\toprule
\multicolumn{1}{r}{ } & \multicolumn{1}{r}{ }& \multicolumn{1}{r}{d\"urfen}& \multicolumn{1}{r}{k\"onnen} & \multicolumn{1}{r}{m\"ussen} & \multicolumn{1}{r}{sollen} \\
\midrule
\multirow{3}{*}{EPOS$_G$ (train)}	 & ep & 1000   & 1000        &  1000   &  1000\\
 & de & 1000   & 1000         &  1000  &   1000\\
 & dy & --   & 1000        &  --   &    -- \\
 \midrule
\multirow{3}{*}{EPOS$_G$ (test)}	 & ep & 98   & 100         & 32   &    100\\
 & de & 98   & 47        &  100     &      100\\
 & dy & --   & 100         &  --    &      --\\ 
\bottomrule
\end{tabular}
}

%
%

\caption{Composition of MPQA, EPOS$_E$, MASC and EPOS$_G$}
\label{tbl:datasets}
\end{table}

\subsection{Experimental settings}

\paragraph{MSC on MPQA using CNN-E$_\textbf{B}$ and CNN-E$_\textbf{U}$, CV}
For MSC we benchmark the CNN approach against the latest state-of-the-art results in Z+. 
We reimplemented their maximum entropy classifier (henceforth, \textit{MaxEnt}) and trained it on their balanced and unbalanced blend of MPQA and EPOS.\footnote{We omit {\em shall} with a small number of instances.} 
As in Z+ we train independent classifiers for each modal verb on their respective training data.\footnote{This holds for all our experiments.}
For evaluation, we perform 5-fold cross validation as in Z+. Each fold for training holds a stratified 80\% section of the MPQA data together with the full EPOS$_E$ data set, and we use the remaining 20\% of MPQA data for testing. We refer to the CNN models trained on the $\pm$balanced versions of this data as CNN-E$_\textbf{B}$ and CNN-E$_\textbf{U}$.

\paragraph{MSC on MASC using CNN-E$_\textbf{B}$ and CNN-E$_\textbf{U}$}
Besides MPQA, we evaluate the CNN on the multi-genre MASC (sub)corpus. 
For comparability with Z+, for training we use one training fold from the previous setting,\footnote{Hence, one 80\% fold of MPQA plus EPOS$_E$. Despite this small difference, we refer to the CNN models as above, as CNN-E$_\textbf{B}$ and CNN-E$_\textbf{U}$.} and evaluate on MASC as test. We analyze the performance of the CNN model overall and on different genre subcorpora (not reported here).

Both English data sets are characterized by modest training set sizes and involve a considerable distributional biases, with high most frequent sense majority baselines (cf.\ Tables \ref{tbl:baselinesmpqa} and \ref{tbl:results_masc}).

\paragraph{MSC on EPOS$_G$ using CNN-G}
In constrast to the English data sets, the German EPOS$_G$ data set provides larger training set sizes of 1000 instances for all modal verbs and senses. This eliminates distributional bias from the data, so that the discriminating power of the classifier model is not masqued by distributional information. 

\subsection{Model variations}

\paragraph{Hyperparameters} Model-specific hyperparameters of the CNN are the number of filters, filter region size, and the depth of the network. We restrict our model to a one-dimensional CNN architecture.

Following the advices in \newcite{ZhangWallace:15}, we 
used following setting: ReLU (rectified linear unit) as activation function, filter region sizes of $3$, $4$, and $5$ with $100$ feature maps each, dropout keep probability of $0.5$, $l_{2}$ regularisation coefficient of $10^{-3}$, number of iterations of $1001$\footnote{We did not perform early stopping.} and mini-batch size of $50$.  
Training is done with the Adam optimisation algorithm \cite{kingma2014adam} with learning rate of $10^{-4}$. 
Filter weights are initialized using Glorot-Bengio strategy \cite{glorot2010understanding}. We experimented with some parameter variations (using nested CV), but found no consistently better results. In all following MSC experiments we thus used this hyperparameter setting for CNN training. 

\paragraph{Word embeddings} In the first and third experimental setting we investigate the impact of static and tuned versions of different word vectors: word2vec \cite{mikolov2013distributed}, dependency-based \cite{levy2014dependency} and randomly initialized embeddings. 

We used publicly available \texttt{word2vec} vectors that were trained on Google News for English\footnote{https://code.google.com/archive/p/word2vec} and various datasets for German \cite{TUD-CS-2014-0973}\footnote{https://www.ukp.tu-darmstadt.de/research/ukp-in-challenges/germeval-2014}, as well as  English dependency-based vectors trained on Wikipedia\footnote{https://levyomer.wordpress.com/2014/04/25/dependency-based-word-embeddings}. The German dependency-based embeddings were trained on the SdeWaC corpus \cite{faass},
parsed with Malt parser. We used 300 dimensions for English embeddings and 100 for German.

For words without a pre-trained vector and in the random initialization setting, each dimension of the random vector was sampled from $\mathcal{U} \sim [-a,a]$ with parameter $a$ picked such that the variance of the uniform distribution equals
the variance of the available pre-trained vectors. 



\paragraph{Baselines}
For MPQA and MASC, the classifiers are compared against strong {\em majority sense baselines, BL$_{maj}$}, due to skewed sense distributions in the training data. 
Further, we compare the CNN results to the reconstructed \textit{MaxEnt} classifier from Z+, trained on the blend of MPQA and EPOS with R\&R's shallow lexical and syntactic path features and the newly designed semantic features of Z+. 

To our knowledge, there is no work on modal sense classification using a neural network. We thus compare our CNN models with a simple, one-layer neural network NN to investigate the impact offered by the more complex CNN architecture.

Input to the NN is the sum of all vectors of the words in the sentence. As for the CNN, we experimented with different types of word vectors.

The \textit{hyperparameter setting} 
for the NN is: ReLU as activation function, $l_{2}$ regularisation coefficient of $10^{-3}$, hidden layer size of $1024$, number of iterations of $3001$, dropout keep probability of $0.5$, and mini-batch size of $50$. Training is again done with the Adam optimisation algorithm \cite{kingma2014adam} with learning rate of $10^{-4}$. Weights are initialized using Glorot-Bengio strategy \cite{glorot2010understanding}.\footnote{This is clearly not shown to be the best hyperparameter setting, as we chose it heuristically without tuning.}

\begin{table}[t]
\centering
\small
\resizebox{0.45\textwidth}{!}{
\begin{tabular}{c|c|c|c|c|c}
\toprule
\multicolumn{1}{r}{CNN-E$_{\text{B}}$}& \multicolumn{1}{r}{can}& \multicolumn{1}{r}{could} & \multicolumn{1}{r}{may} & \multicolumn{1}{r}{must} & \multicolumn{1}{r}{should}\\
\midrule

    w2v-static & 65.02 & 51.67 & \textbf{93.57} & \textbf{93.82}  & \textbf{90.77}\\ 
    w2v-tuned & 63.73 & 54.17  & \textbf{93.57} & \textbf{93.82} &  \textbf{90.77}\\ 
    dep-static & \textbf{65.78} & 56.67 & \textbf{93.57}  & \textbf{93.82} & \textbf{90.77}\\ 
    dep-tuned & 59.89 & \textbf{67.50}  & \textbf{93.57} & 93.29 & 90.42\\
    rand-static & 63.99 & 46.67 & \textbf{93.57} & 92.79 & \textbf{90.77}\\
    rand-tuned & 64.50 & 48.33 & \textbf{93.57} & 92.79  & \textbf{90.77}\\
\bottomrule
\end{tabular}
}

\vspace{0.1cm}
\centering
\small
\resizebox{0.45\textwidth}{!}{
\begin{tabular}{c|c|c|c|c|c}
\toprule
\multicolumn{1}{r}{CNN-E$_{\text{U}}$}& \multicolumn{1}{r}{can}& \multicolumn{1}{r}{could} & \multicolumn{1}{r}{may} & \multicolumn{1}{r}{must} & \multicolumn{1}{r}{should}\\
\midrule

    w2v-static & 70.10 & 65.27 & \textbf{93.49} & \textbf{94.97}  & \textbf{90.59}\\ 
    w2v-tuned & 70.62 & 66.10  & \textbf{93.49} & \textbf{94.97} &  \textbf{90.59}\\ 
    dep-static & 69.85 & 65.27 & \textbf{93.49}  & 94.46 & \textbf{90.59}\\ 
    dep-tuned & 69.59 & \textbf{66.55}  & \textbf{93.49} & 93.95 & \textbf{90.59}\\
    rand-static & 70.36 & 64.45 & \textbf{93.49} & 93.45 & \textbf{90.59}\\
    rand-tuned & \textbf{70.87} & 64.86 & \textbf{93.49} & 93.45  & \textbf{90.59}\\
\bottomrule
\end{tabular}
}

\vspace{0.1cm}
\resizebox{0.45\textwidth}{1.6cm}{
\begin{tabular}{c|c|c|c|c}
\toprule
\multicolumn{1}{r}{CNN-G} & \multicolumn{1}{r}{d\"urfen}& \multicolumn{1}{r}{k\"onnen}& \multicolumn{1}{r}{m\"ussen} & \multicolumn{1}{r}{sollen}\\
\midrule
    w2v-static & 91.92 & 68.82 & 77.61 & 71.64 \\ 
    w2v-tuned & \textbf{99.49}  & 74.09 & 83.58 & 72.14\\
    dep-static & 91.92 & 63.56 & 75.37& 73.13\\ 
    dep-tuned & 97.47 & 73.28  & 82.83 & \textbf{74.63}\\
    rand-static & 96.46  & 77.33 & 81.34 &  74.13\\
    rand-tuned & 98.48  & \textbf{78.95} & \textbf{85.07}& 73.63\\
\bottomrule
\end{tabular}
}

\caption{CV accuracy for CNN-E$_{\text{B}}$, CNN-E$_{\text{U}}$, test accuracy for CNN-G, with different input representations.}
\label{tbl:wordvectors}
\end{table}

\subsection{Results}

\subsubsection*{English}

In Table \ref{tbl:wordvectors} we report results for CNN-E$_{\text{B}}$ and CNN-E$_{\text{U}}$ with diverse input representations. For balanced training, dependency based vectors yield the best (\textit{can}, \textit{could}) or equally good results (\textit{may}, \textit{must}, \textit{should}). \textit{Could} is the only case with large performance differences
depending on the choice of embeddings. For \textit{can} and \textit{could} choosing either static or tuned versions of vectors is beneficial. With unbalanced training, dependency-based vectors are outperformed by \texttt{word2vec} for \textit{must} and by randomly initialized vectors for \textit{can}. Large differences in the results for \textit{could} w.r.t.\ the choice of embeddings, are no longer present.

In Table \ref{tbl:baselinesmpqa} we report overall results for CNN-E$_{\text{B}}$ and CNN-E$_{\text{U}}$ on MPQA compared to the baselines. As representations for the NN and CNN we selected, for each modal verb, the embedding type that yielded the best results (Table \ref{tbl:wordvectors})\footnote{For NN the impact of word vectors was investigated as well.}.

For each training data set, scores of the CNN which are significantly better\footnote{By conducting the mid-p-value McNemar test \protect\cite{fagerland2013mcnemar} with p \textless 0.05.} than the next lower score among the baselines  are \underline{underlined}. If CNN does not yield the best results, significance between the baseline with the best score and CNN is reported. $\overline{\text{Overlining}}$ is used if CNN with unbalanced training performs significantly better than CNN with balanced training, and vice versa. 

With balanced training, CNN outperforms all baselines for every modal verb and in terms of micro average. However, differences between CNN and \textit{MaxEnt} are significant only for \textit{can}, \textit{could} and micro average. Moving to unbalanced training, CNN has difficulties beating the baselines (cf.\ \textit{may}, \textit{should}), but yields the best micro average. Unbalanced training for CNN outperforms balanced training in terms of micro averages, however the difference is not significant.

Table \ref{tbl:results_masc} summarizes the evaluation of CNN-E$_{\text{B}}$ and CNN-E$_{\text{U}}$ on the MASC corpus. Note that CNN with unbalanced training, CNN-E$_{\text{U}}$, does not have enough generalization capability when applied to different genres. This behavior coincides with 
changes of the predominant sense between training and test. CNN-E$_{\text{U}}$,  as well as \textit{MaxEnt}, 
is highly sensitive to such distributional changes. Even though balanced training for CNN leads to a slightly worse micro average when evaluated on MPQA, on MASC CNN--E$_{\text{B}}$ yields a +3pp gain in micro average compared to unbalanced training.\footnote{In contrast to {\em MaxEnt}, which does not profit from balanced training.}

In sum, our evaluation shows that the CNN model is able to outperform strong baselines in most configurations. Balanced training shows more consistent results beyond the baselines and is competitive with unbalanced training, without significant difference except for {\em can}. 
In view of genre differences in MASC, the CNN--E$_{\text{B}}$ model is more robust against sense changes, and yields overall better results. The strong behaviour on balanced training data shows that the CNN model is able to learn meaningful structure from the data.

\begin{table}[t]
\centering
\small
\resizebox{0.48\textwidth}{!}{
\begin{tabular}{@{}ccccccc@{}}
\toprule
            & can & could & may & must & should & micro\\
\midrule
BL$_{rand}$ & 33.33 & 33.33 & 50.00 & 50.00 & 50.00 & 41.49\\
MaxEnt & 59.64 & 61.25 & 92.14 & 87.60 & 90.11 & 74.88\\
NN & 56.01 & 55.42 & 90.00 & 75.24 & 88.68 & 69.74 \\ 
\midrule
CNN-E$_{\text{B}}$ & \underline{\textbf{65.78}} & \underline{\textbf{67.50}} & \textbf{93.57} & \textbf{93.82} & \textbf{90.77} & \underline{\textbf{79.29}}\\
\bottomrule
\end{tabular}
}

\vspace{0.1cm}
\centering
\small
\resizebox{0.48\textwidth}{!}{
\begin{tabular}{@{}ccccccc@{}}
\toprule
            & can & could & may & must & should & micro\\
\midrule
BL$_{maj}$ & 69.92 & 65.00 & 93.57 & 94.32 & 90.81 & 80.18   \\
MaxEnt & 64.76 & 63.33 & 92.14 & 92.78 & \textbf{91.48} & 78.01\\
NN & 67.29 & 66.08 & \underline{\textbf{94.23}} & 86.37 & 90.96 & 77.93 \\ 
\midrule
CNN-E$_{\text{U}}$ & $\overline{\textbf{70.87}}$ & \textbf{66.55} & 93.49 & \textbf{94.97} & 90.59 & \textbf{80.74}\\
\bottomrule
\end{tabular}
}

\caption{\footnotesize Comparison of CV accuracies on MPQA of CNN-E$_{\text{B}}$ (upper table) and CNN-E$_{\text{U}}$ (lower table) with baselines. } 
\label{tbl:baselinesmpqa}
\end{table}

\begin{table}[t]
\centering
\small
\resizebox{0.48\textwidth}{!}{
\begin{tabular}{@{}ccccccc@{}}
\toprule
& can & could & may & must & should & micro\\
\midrule
BL$_{rand}$ & 33.52 & 33.82& 48.67 & 46.87 & 46.01&  38.63 \\
MaxEnt & 66.74 & 62.86 & {\bf 87.83} &83.33 & 84.06 & 72.25 \\
\midrule
CNN-E$_{\text{B}}$ & {\bf 80.46} &  {\bf 64.48} & 86.69 & {\bf84.72} & {\bf88.84} &  {\bf79.33} \\
\bottomrule
\end{tabular}
}

\vspace{0.1cm}
\resizebox{0.48\textwidth}{!}{
\begin{tabular}{@{}ccccccc@{}}
\toprule
& can & could & may & must & should & micro\\
\midrule
BL$_{maj}$ & {\bf81.61}        & 35.04      & 82.51 			& 79. 86                 & 89.24                & 72.86   \\
MaxEnt & 73.17  & {\bf 55.34}& {\bf 87.45} & 86.11                           & {\bf 89.64}  			  & 74.41\\
\midrule
CNN-E$_{\text{U}}$ & 81.03 & 49.15 & 86.31 & {\bf86.80} & 89.24 & {\bf76.49} \\
\bottomrule
\end{tabular}
}
\caption{Accuracies on MASC dataset of classifiers trained on MPQA+EPOS$_E$.} 
\label{tbl:results_masc}

\end{table}

\subsubsection*{German}

In Table \ref{tbl:wordvectors} we report results for CNN-G with diverse input representations. Reasons for the slightly weaker performance of dependency-based vectors compared to word2vec (1-2 pp.) can be seen in the smaller size of the training corpus, and possibly greater noise due to parsing errors.

In Table \ref{tbl:results_german} we report overall results for CNN-G compared to the NN baseline.\footnote{We did not construct a MaxEnt classifier for German. For NN and CNN-G we chose the best performing embedding types per modal verb.} The CNN outperforms both baselines by large margins, per modal verb and in terms of micro average.
Given we employed perfectly balanced training data, the classifier performances reflect their ability to learn characteristic information for the classes. 
 Indeed, the NN has great difficulties distinguishing the senses for {\em k\"onnen} (3 senses) and {\em sollen}, and is outperformed by CNN-G by +35.6 and +24.4 pp.\ gains. The confusion matrices for CNN-G show a clear separation of these classes, in contrast to the NN.
 
 While German is a more difficult language than English due to its syntactic properties (word order, degree of inflection), CNN-G reaches overall higher performance levels compared to English, especially for difficult cases.\footnote{Clearly, we cannot draw any strict comparison here.}
One reason can be the morphological distinction between indicative and subjunctive (Konjunktiv), which --  in interaction with tense and other factors -- can ease the distinction of epistemic vs.\ deontic/dynamic sense. For {\em sollen} this morphological division is masqued, and this can explain the weaker results compared to other binary classes.
Generally, CNN-G profits from larger and perfectly balanced training data.

\begin{table}[t]
\centering
\small
\resizebox{0.48\textwidth}{!}{
\begin{tabular}{@{}cccccc@{}}
\toprule
&d\"urfen& k\"onnen & m\"ussen & sollen  & micro\\
\midrule
BL$_{rand}$ & 50.00 & 33.33 & 50.00 & 50.00 & 39.10\\
NN & 77.73 & 43.32 & 73.88 &  50.25 & 57.69\\ 
\midrule
CNN-G & \underline{{\bf99.49}} & \underline{{\bf78.95}} & \underline{{\bf85.07}} & \underline{{\bf74.63}} & \underline{{\bf 84.10}}\\
\bottomrule
\end{tabular}
}
\caption{Average accuracy on EPOS$_G$.} 
\label{tbl:results_german}
\end{table}

\subsection{Semantic feature detectors}

Z+ provided a thorough analysis of the impact of 
semantic features by ablating individual feature groups. Their ablation 
analysis confirmed that feature groups relating to tense and aspect 
of the embedded verb, negation, abstractness of the subject and semantic features of the embedded verb yield significant effects on classification performance. 

For \textit{must}, Z+ found clear patterns for the occurrence of specific features and the ability to properly classify a specific sense. However, they did not identify precise features that differentiate epistemic and dynamic readings with \textit{can}. We speficically investigated whether the learned filters for {\em must} can be related 
to the  semantic categories Z+ found to be important for distinguishing its senses.
In addition, we investigated whether the CNN is able to capture unattested features that differentiate epistemic and dynamic readings with \textit{can}.

For every modal verb and every filter, we sort sentences in the training data by the maximum value obtained by applying \textit{1-max pooling} to the feature map acquired by applying the respective filter to a sentence. 
For each filter and each of the top-ranked 15 sentences, we extract the
ngram that corresponds to the maximum value w.r.t.\ the filter, i.e.\ the argmax of the feature map. The ngram vector is the sum of all vectors of words in the ngram. The obtained ngram vectors
were plotted using the t-SNE algorithm \cite{van2008visualizing} and textually displayed with their surrounding context.


For \textit{must} we found many feature detectors that relate to observations in Z+. Many filters detect past (\textit{\textbf{you must have been} out last night}; ep)  vs.\ non-past (\textit{\textbf{we must make} further efforts}; de) 
and a dynamic event (\textit{\textbf{we must develop} a policy}; de) vs.\ stative (\textit{\textbf{you must think} me a perfect fool}; ep) reading of the embedded verb. Among others the feature detectors capture passive constructions (\textit{actual steps \textbf{must be taken}}; de)
and negation (\textit{\textbf{we must not} fear}; de). Some filters were trained to capture domain vocabulary which intuitively goes along with deontic sense ({\em European parliament; present regulation; fisheries policy}). 
One filter captures telic clauses ({\em to address these problems; to prevent both forum; to exert maximum influence}), identifying deontic sense. 
Novel features not considered in 
Z+ are discourse markers ({\em but; and (then)})
that 
correlate with deontic sense. All in all, 
the CNN learns meaningful features that are known to be important for differentiating senses for \textit{must}, and in contrast to manual feature design, it detects relevant unattested  features by itself.

For \textit{can} many filters 
recognise accomplishments which go along with dynamic sense, e.g. \textit{You can do it/make it to NY}. Others detect  words indicating \textit{possibility} (ep), negation (de), discourse markers, animate subject (de and dy), passive construction (de and dy).   However, without a systematic classification of these features it remains unclear how important they are for differentiating the senses of \textit{can}. Also, similar to Z+ we did not find clear-cut features that recognize epistemic sense.

We performed a corresponding analysis of feature maps for German, following the same extraction procedure. We found the typical state (ep) vs.\ event (de) contrast for the embedded verb, negation and tense, and again previously unattested factors such as discourse relation markers\footnote{For reasons of space we provide translations to English.} ({\em but; without; thereby; in order to} (dy)). For German we identified various indicators for epistemic sense (for {\em m\"ussen} and {\em k\"onnen}): attitude predicates ({\em believe, not know; tell me; have an idea, be afraid}), adverbials ({\em  possibly}), conditionals ({\em if}); counterfactual and negative polarity contexts ({\em not be the case; how; ever}).  Further detectors for epistemic sense are abstract subjects: placeholders for propositions ({\em it}), abstract concepts ({\em idea; music; grades; application}); indefinite subjects ({\em one}). We find a tendency for 1st or 2nd person subjects to co-occur with de/dy and 3rd person pronouns with ep. For {\em k\"onnen} (dy) we find achievements ({\em present report;  move mountains; find compromise}). For deontic readings, next to negation with 1st and 2nd person we find typical verb-object combinations for actions that can be granted: {\em use telephone; communicate with third parties}.

We extracted statistics about the distance of the extracted ngrams from the modal verb  (distance overall; to the left/right and ngrams starting with the modal). There are no greater overall distances for German compared to English. However, for German we find significantly more ngrams that include the modal verb, especially for epistemic readings of {\em k\"onnen, m\"ussen, d\"urfen} that clearly mark subjunctive mood, whereas for {\em sollen}, with ambiguous forms for subjunctive and past tense, no such tendency is observed. Thus, the feature maps identify subjunctive marking (in conjunction with other factors) as relevant for classifying epistemic sense, whereas for {\em sollen} the lack of this indicator goes along with lower performance. Finally, we observe, for English and German, strikingly larger distances to the left of the modal verb for epistemic readings compared to non-epistemic readings.
This can be traced back to indicators in the wider left-embedding context: embedding predicates, subjects, if clauses, etc. 
\section{Word sense disambiguation}
Next to modal sense classification, we evaluate our CNN model in a classical WSD task. As benchmark corpus we chose the {\em SensEval-3 lexical sample} data set \cite{mihalcea2004sel}, which was recently applied in \newcite{rothe-schutze:2015:ACL-IJCNLP} (henceforth R\&S) and \newcite{taghipour2015semi}, using sense-specific embeddings and a NN architecture, respectively (cf.\ Section 2). 

%
%
%
%

The training data size for the 57 target word types ranges from 14 to 263 instances. Sense labels of test instances of a given target word are predicted using the CNN model trained on the training instances for the respective word type.
\footnote{Training instances in the {\em SensEval-3} dataset can have more than one sense label. For training we randomly picked one of possible labels. Instances which contain more than one marked target word were omitted.}
We set the CNN hyperparameters to be the same as for MSC, except for mini-batch size and region sizes. Since the training data for some words is below $50$ instances, mini-batch size was set to $10$. For tuning of the region sizes, we split the training data for each word (80:20 for training and validation) and used static \texttt{word2vec} for the input representation. Among $\{(1, 2, 3), (2, 3, 4), (3, 4, 5), (4, 5, 6), (5, 6, 7)\}$ the best results were obtained for $(5,6,7)$.\footnote{However, the differences in the results were minor.}

The final hyperparameter setting was used to investigate the impact of representations. Among \texttt{word2vec}, dependency-based and randomly initialised, \texttt{word2vec} performed the best, the tuned version being slightly better than static vectors. We report results for tuned \texttt{word2vec} vectors.

We compare our results 
to the results R\&S 
obtained when using only sense-specific embeddings. These are not the state-of-the-art WSD results they obtain with additional features, namely POS tags of words in a small window around the target word, their discrete representation and local collocations. For sentence representation, R\&S
used every word in the target word sentence.
For sense prediction, they used the following feature vectors that are fed into a linear SVM classifier: 

\vspace*{-4mm}
{\small
\begin{align*}
\text{S-cosine} &= \langle \text{cos}(c, s^{(1)} ), \hdots, \text{cos}(c, s^{(k)}) \rangle \,,\\
\text{S-product} &= \langle c_{1} s^{(1)}_{1}, \hdots, c_{n} s_{n}^{(1)} , \hdots, c_{1} s_{1}^{(k)}, \hdots, c_{n}s_{n}^{(k)} \rangle\,,\\
\text{S-raw} &= \langle c_{1}, \hdots, c_{n}, \hdots, s_{1}^{(k)} , \hdots, s_{n}^{(k)} \rangle\,,
\end{align*}\par}
\vspace*{-3mm}

\noindent
where $w$ is a target word with $k$ senses, $c$ is the centroid defined as the sum of all \texttt{word2vec} vectors of words in the sentence and $s^{(j)}$ is the embedding of the $j$-th synset of $w$.\footnote{Obtained using the \textit{AutoExtend} method of R\&S.} 
They propose a variant of the {\em S-prod} feature vector, {\em S$_{naive}$-prod}, for which the synset embeddings are the sum of the \texttt{word2vec} vectors of all words in that sysnet. 


\begin{table}[t]
\centering
\small
\resizebox{0.23\textheight}{!}{
\begin{tabular}{@{}cccc@{}}
\toprule
S$_{naive}$-prod & 62.20 & S-prod & 64.30 \\
S-cosine & 60.50 & S-raw & 63.10 \\ 
\midrule
& & CNN & \textbf{66.50}\\
\bottomrule
\end{tabular}
}
\caption{WSD accuracy on {\em SensEval-3} dataset.} 
\label{tbl:results_wsd}
\end{table}

The results are summarised in Table \ref{tbl:results_wsd}. The CNN model compares favorably to the competitor models of R\&S using \textit{AutoExtend} embeddings for WSD. It achieves slightly higher results without explicitly marking  the target word, whereas the \textit{AutoExtend} embeddings encode much richer information: what is the target word, how many possible sense it has, and knowledge-intense sense embeddings for each of its synsets. The CNN  is able to compete with the rich {\em AutoExtend} model, and future work needs to investigate whether -- similar to the S-product setting in R\&S -- the CNN model can 
achieve competitive state-of-the-art results by 
 incorporating features corresponding to those of the 
IMS system of
 \newcite{zhong2010makes}.





\section{Conclusion and future work}

We presented an account for multilingual modal sense classification using a CNN architecture. We apply the same architecture in a standard WSD task and achieve competitive results compared to a system using richer embedding information.

Our one-layer CNN architecture outperforms strong baselines and prior art for MSC in English, including a NN and MaxEnt model, and proves particularly robust in cross-genre classification. 

We applied the CNN model to German, on a data set of modest size, obtained using cross-lingual projection techniques. The CNN-G classifier outperforms a NN model by large margins. 

Our approach can be easily generalized to novel languages without tedious and resource-intensive feature engineering.
 Through analysis of learned feature maps we gave evidence that the CNN learns both known and novel  features for MSC.

The  attractiveness of the CNN framework lies in
its ability to learn (semantic) features from flexible window regions  without syntactic processing, and the ensuing robustness on difficult text genres and its ease in generalizing to novel languages.





\section*{Acknowledgments}
We thank Mengfei Zhou for her support with the German corpus construction. This work has been supported by the German Research Foundation as part of the Research Training Group "Adaptive Preparation of Information from Heterogeneous Sources" (AIPHES) under grant No. GRK 1994/1.

\bibliography{references}
\bibliographystyle{acl2016}




\end{document}